# Genetic Algorithm-Based Coordination and Optimization Model for "Generation-Grid-Load-Storage" in Active Distribution Networks

Jinlu Zhang[a,1], Fujian Chi[a] , Tianhan Ling[b] , Yulong He[a] , Kejia Zhang[a], Hongxing Lv[a] , Sheng Wang[a]

[a]*State Grid Tianjin Binhai Electric Power Supply Company, Tianjin, 300450, Tianjin, China*

[b]*State Grid Tianjin Electric Power Company, Tianjin, 300000, Tianjin, China*

**Abstract.** Create an optimization framework that combines fuzzy logic and genetic algorithms for risk assessment and coordination of generation, grid connection, load, and energy storage facilities in active distribution networks. In order to capture the system uncertainties caused by weather factors and dynamic user demands, this project uses fuzzy set representation to model renewable energy generation, load curves, and market prices. In the evolutionary process, genetic algorithms introduce fuzzy elements into system management, adjusting thru penalty factors to improve stability and seeking better scheduling or power dispatch solutions. At the same time, by penalizing uncertainty and constraint violations, the fitness function of the hybrid model aims to optimize the expected operational costs; it can produce feasible and economical results under unknown parameter conditions. Under conditions with a large supply of renewable energy and installed energy storage systems, simulation results for the IEEE-69 power system indicate that the fuzzy genetic algorithm strategy effectively reduces technical constraints compared to deterministic optimization schemes. The total investment remains at a similar level. Other data indicate that thru fuzzy reasoning optimization, unreasonable choices or impossible situations are avoided during the network adaptation process. This framework-based technical approach helps to construct an effective computational scheme that is aware of uncertainty in its outputs. This will provide a scientific basis for the uncertainty assessment in distribution network planning.

**Keywords.** Computational Optimization, Fuzzy Logic, Genetic Algorithm, Active Distribution Network, Uncertainty Quantification, Power System Security

## 1. Introduction

Due to the increased demand for renewable energy in active distribution networks, the economics and fundamental technical assumptions of power system operation have changed [1]. In some local grid systems, the proportion of photovoltaic and wind power generation approaches or exceeds 40% of the total installed capacity, leading to an increase in the scale and frequency of output instability [2]. According to an empirical

---

analysis of operational log data from major distribution systems in Europe and Asia, the planned outages and voltage deviations at nodes and regions have significantly increased compared to the past [3]. Due to certain new energy applications, such as electric vehicles and variable loads, accurately predicting network operating conditions and market information has become more difficult, leading to increased volatility in demand patterns [4].

Most optimization frameworks used to coordinate generation, grid, load, and energy storage in ADN employ deterministic or scenario models to assume that system parameters are fully available [5]. In practical system applications, traditional mathematical programming techniques are more likely to overestimate system risks[6]. Simulations of large-scale demonstration projects indicate that deterministic models will lead to poor operational planning, with approximately a 50% increase in constraint violations[7]. In order to better measure and address operational uncertainties in active distribution systems, it is necessary to optimize the system architecture to eliminate the discrepancies between model assumptions and the actual system conditions[8].

A new hybrid coordination and optimization method combines fuzzy logic uncertainty modeling with complex genetic algorithms to overcome the shortcomings of the aforementioned models. The proposed method constructs fuzzy set representations of the uncertain behaviors of renewable energy generation, load, and price signals. A robust, iterative genetic algorithm search is also used to handle uncertainty. By combining strict uncertainty with intelligent optimization methods, the robustness and economy of the system during operation in ADM can be improved. A feasible solution is provided to address the current issues faced by the smart grid.

## 2. Fuzzy Logic-Based Uncertainty Modeling

### *2.1. Fuzzy Set Representation of Uncertain Variables*

Fuzzy set theory can provide strong support for modeling the fluctuations of active distribution network. Due to multi-scale uncertainties, active distribution networks exhibit volatility in terms of generation, market, and demand factors[9]. Environmental factors such as weather can cause intermittent failures and low accuracy in wind turbines and photovoltaic systems. In the context of ADN, renewable energy output is integrated thru fuzzy logic; the membership function shows various possible outcomes based on past data and prediction errors. Due to the identical load demand patterns of network nodes, they can be considered an uncertain parameter, allowing for randomness or behavioral deviations. Most scholars believe that the current instantaneous electricity price is an uncertain economic indicator due to external shocks or changes in the demand-supply relationship. Wind and solar power generation are typical membership function curves of renewable energy. Figure 1 shows the envelope, which is usually represented using trapezoidal or Gaussian functions. Uncertainty representation can be flexibly expressed and is sufficiently concise in subsequent optimization objectives.

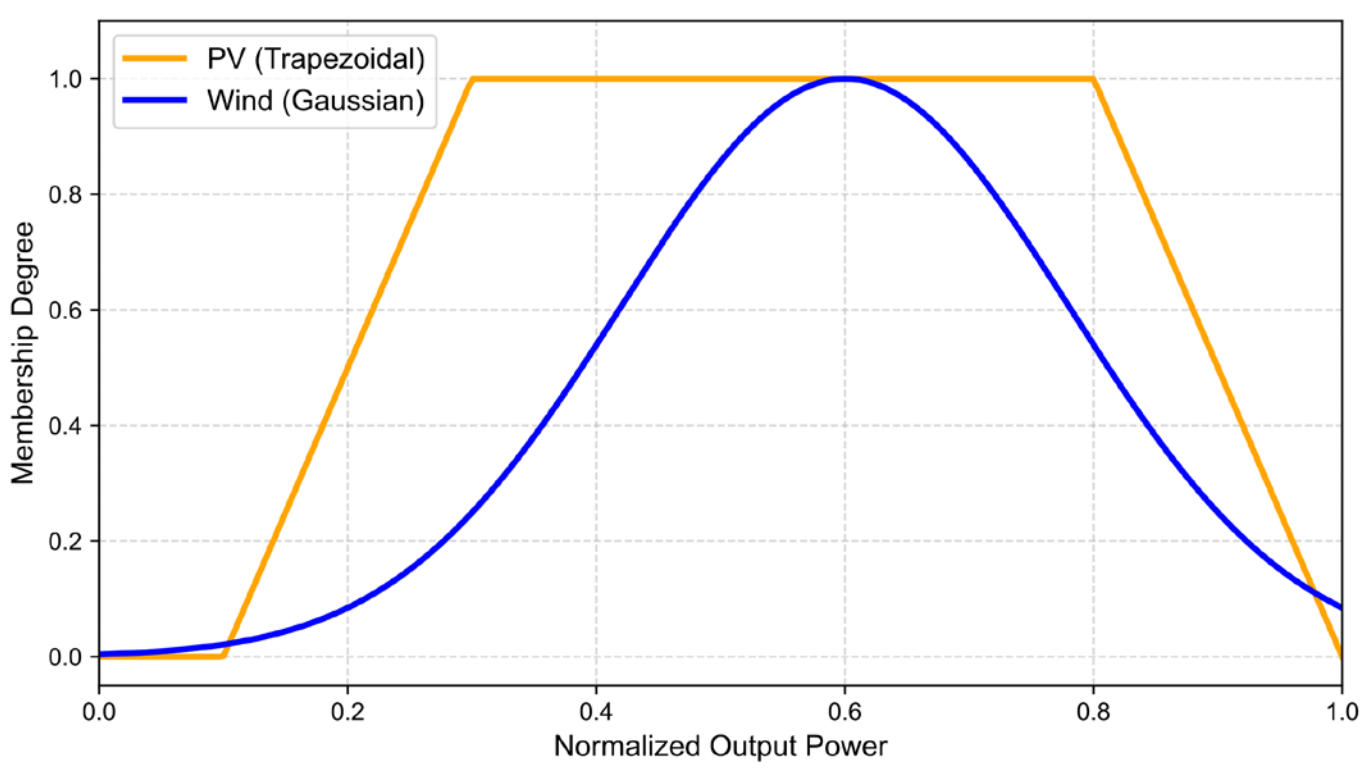

**Figure 1.** Membership function curves of photovoltaic and wind generation output.

Fuzzy membership functions are derived from empirical data and operational risk characteristics, rather than being randomly chosen. For example, Gaussian membership functions tend to use symmetry in wind density predictions; for solar radiation predictions, trapezoidal membership functions with discontinuities at sunrise and sunset are relatively reasonable. By calibrating the above functions based on historical system data and obtaining approximate estimates; quickly adjust the time when seasonal changes or other sporadic events occur. Thru data-driven optimization, fuzzy representations associated with each uncertain variable can be accurately captured, while also capturing regular variations and extreme events. By incorporating these statistically guided membership functions into modeling methods, theoretical credibility and practical effectiveness can be provided for real-time scheduling applications of large-scale renewable energy-driven distribution networks.

*2.2. Defuzzification and Confidence-Governed Constraint Modeling*

For operational integration, fuzzy variables must be translated into actionable crisp values without discarding the valuable uncertainty information inherent to their nature. Among available defuzzification mechanisms, the centroid method is most effective, as it computes the expected value by considering the full distribution of possibilities. Specifically, the centroid $P_G^*(t)$ of the fuzzy output variable is given by:

$$P_G^*(t)=\frac{\int_{P_{\min}}^{P_{\max}} p\cdot\mu_{\tilde{P}_C}(p)dp}{\int_{P_{\min}}^{P_{\max}} \mu_{\tilde{P}_C}(p)dp} \tag{1}$$

This continuous weighted integration assures that solutions remain robust even in the presence of multimodal and asymmetric uncertainty. For certain high-stakes decision points-where operational conservativity is preferred-the maximum membership principle may be adopted instead, choosing $p$ at which $\mu_{P_C}(p)$ attains its maximum. However, this heuristic can introduce bias and may underutilize information embedded in the uncertainty landscape, as it is sensitive to outliers and fails to reflect the spread of possible situations [11].

A further essential step in modeling is the incorporation of confidence-governed probabilistic constraints. Rather than enforcing deterministic feasibility, system operation is required to satisfy technological and safety thresholds (such as bus voltages $V_j$ and line currents $I_l$ ) with high probability-conventionally set at 95%. This formalism is embodied in expressions such as:

$$\mathbb{P}\left(V_j^{\min} \leq V_j(t) \leq V_j^{\max}\right) \geq 0.95 \tag{2}$$

and

$$\mathbb{P}(|I_l(t)| \leq I_l^{\max}) \geq 0.95 \tag{3}$$

The calculation of probabilities is based on defuzzification of scenarios derived from fuzzy sets. This probabilistic constraint handling meets operational safety requirements, meaning that all acceptable solutions are robust for most system implementations within the model's fuzzy set [12]. A flexible yet solid theoretical foundation has already been established for constraint satisfaction. This constraint satisfaction can fully manage the intermittency of renewable energy supply and demand fluctuations under feasible and mathematically rigorous conditions.

## 3. Hybrid Fuzzy-GA Optimization Algorithm

### *3.1. Algorithm Integration Framework*

After the classic genetic algorithm operations (such as selection, crossover, and mutation) and the initial generation of the chromosome population, each potential competitor can only be tested thru non-deterministic performance metrics and fuzzy logic evaluation areas. By interpreting the degree of constraint satisfaction thru fuzzy constraints and probabilistic boundaries, feedback is returned to dynamically adjust the membership function parameters and defuzzification weights, thereby improving the learning of subsequent generations. Figure 2 shows the process of information transmission and feedback between the genetic algorithm kernel and the fuzzy set engine.

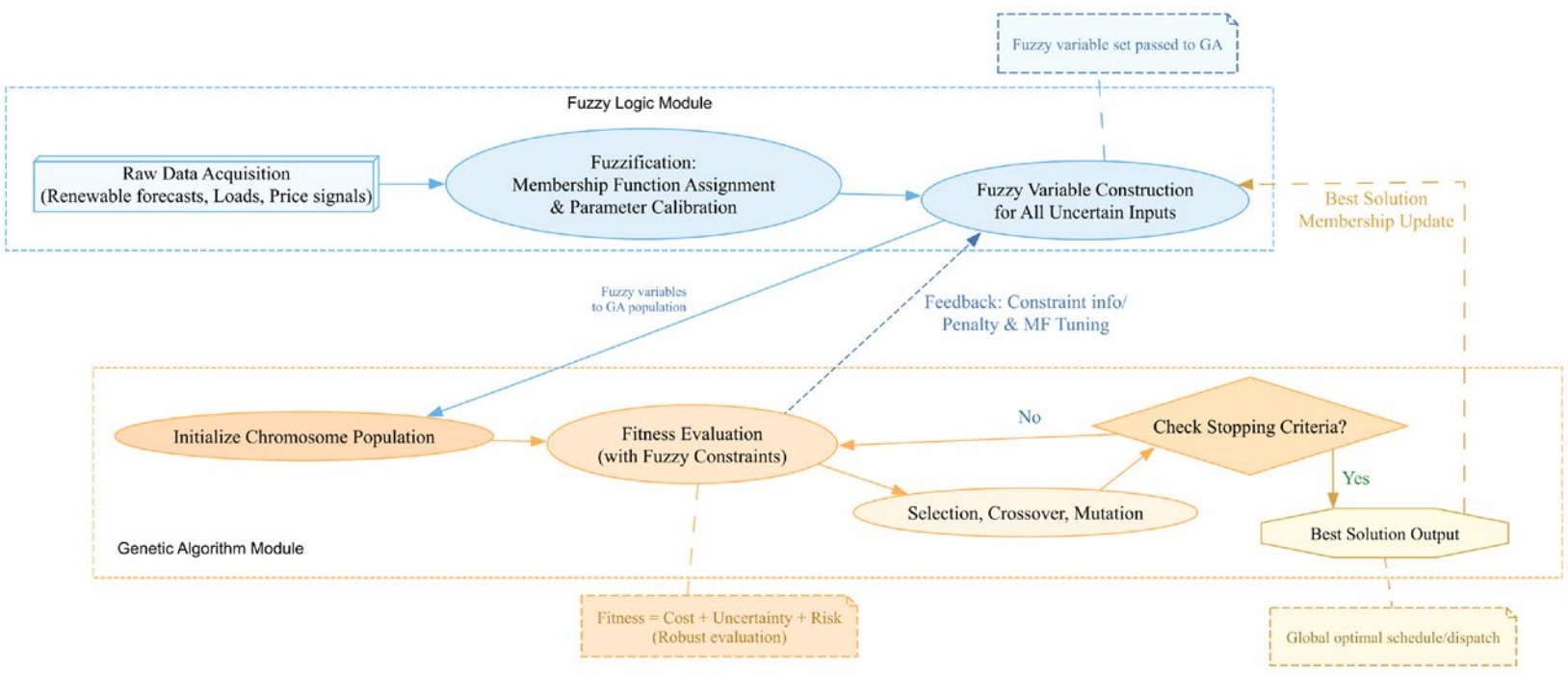

**Figure 2.** Fuzzy-GA algorithm flowchart.

In this design, the uncertainty-based preprocessing step uses fuzzy logic modules to convert raw input data (such as renewable energy production forecasts, load forecasting models, and price signals) into organized fuzzy variable structures [13]. Fuzzy sets are used to encode the uncertainty and fuzziness in the operation of the ADN. Fuzzy sets are sent to the knowledge interface of the GA module, which then performs a global search and identifies the results of constraint avoidance. The bidirectional structure is very natural.

*3.2. Robust Fitness Function Design*

To guarantee both economic efficiency and operational reliability under stochastic disturbance, the hybrid Fuzzy-GA employs a multi-component fitness function of substantial complexity [14]. The total objective function $F$ is constructed as the sum of the weighted economic cost, an uncertainty-induced penalty, and terms reflecting constraint violation risks.

$$F = C_{\mathrm{op}} + \lambda_u U\left(\tilde{P}_G, \tilde{D}, \tilde{\pi}\right) + \lambda_c \Xi(V, I) \tag{4}$$

Here, the term $C_{\mathrm{op}}$ aggregates the conventional operational expenditures, calculated as

$$C_{\mathrm{op}} = \sum_{t=1}^{T} \left(\sum_{g=1}^{G} c_g P_g(t) + \sum_{s=1}^{S} c_s Q_s(t) + \pi(t) \cdot \Delta L(t)\right) \tag{5}$$

where $P_g(t)$ and $Q_s(t)$ denote real and reactive power dispatches across generators and storage units, and $\Delta L(t)$ is the system demand mismatch.

The uncertainty penalty $U$ measures the expected deviation from nominal operation due to input fuzziness and is regulated by the factor $\lambda_u$, ensuring that excessive risk levels downgrade fitness even if costs are conventionally minimized.

Simultaneously, the constraint violation term $\Xi$ is defined as the weighted sum of probabilityadjusted violation magnitudes for all nodes and branches,

$$\Xi(V, I) = \sum_j \chi_j \max\left(0, \mathbb{E}\left[V_j(t)\right] - V_j^{\max}\right) + \sum_\ell \psi_\ell \max(0, |\mathbb{E}[I_\ell(t)]| - I_\ell^{\max}) \tag{6}$$

where penalty coefficients $\chi_j$ and $\psi_\ell$ can be adaptively scheduled according to operational criticality or regulatory constraints [15]. This intertwined, multi-modal objective structure ensures that the final operational plan is robust against both economic and technical risks, maintaining feasibility even in the tail ends of the uncertainty distribution.

*3.3. Sensitivity Analysis of Fuzzy Membership Functions*

In order to improve the stability and effectiveness of the fuzzy genetic algorithm optimization method in active distribution networks, it is necessary to improve the fuzzy parameterization method. In the system model, the uncertainty in renewable energy output or load changes is directly influenced by altering the width and form of the membership function. Systematically adjust the width of the membership function and conduct sensitivity experiments to evaluate its impact on operational costs and the violation of technical constraints.

In smaller membership function settings, the optimizer may be constrained to overly optimistic predictions within a limited uncertainty range. Simulation results indicate that, in this case, operational costs are generally lower, but the risk of violating constraints is higher. The network may first experience voltage or current deviation boundary faults because it is less resilient to prediction errors and unexpected disturbances.

However, due to its broader functional range, it can adapt to more types of uncertainty. Therefore, the occurrence of constraints is reduced, thereby improving the overall reliability of the scheduling solution. However, due to the cautious scheduling settings preferred for optimization, this will increase operational and maintenance costs. Analysis indicates that, in terms of cost-effectiveness and technical risk, an appropriate range of member functions is suitable. Accurately and flexibly selecting membership functions is essential to ensure the stable operation of the active distribution network under real conditions.

## 4. Case Study on Uncertainty-Aware ADN Operation

### *4.1. Test System Description*

Due to its complex topology and numerous nodes with different characteristics, the IEEE-69 bus system is an important benchmark in the study of optimal operation of distributed generation in distribution systems. Evidence is provided for the effectiveness and applicability of the proposed fuzzy genetic algorithm coordination algorithm. Figure 3 shows the comparison results of total operating costs and technical constraint violation rates between GA and Fuzzy-GA methods in the IEEE 69-bus system active distribution network under low, medium, and high uncertainty conditions. The simulation results indicate that the constraint violation rate of the fuzzy genetic algorithm framework is up to 40% lower than that of the traditional genetic algorithm method, but the overall system cost of both methods slightly increases under higher operational uncertainty conditions. All test cases demonstrated excellent performance, confirming that the application of fuzzy logic-based uncertainty modeling in the optimization process is feasible both economically and operationally.

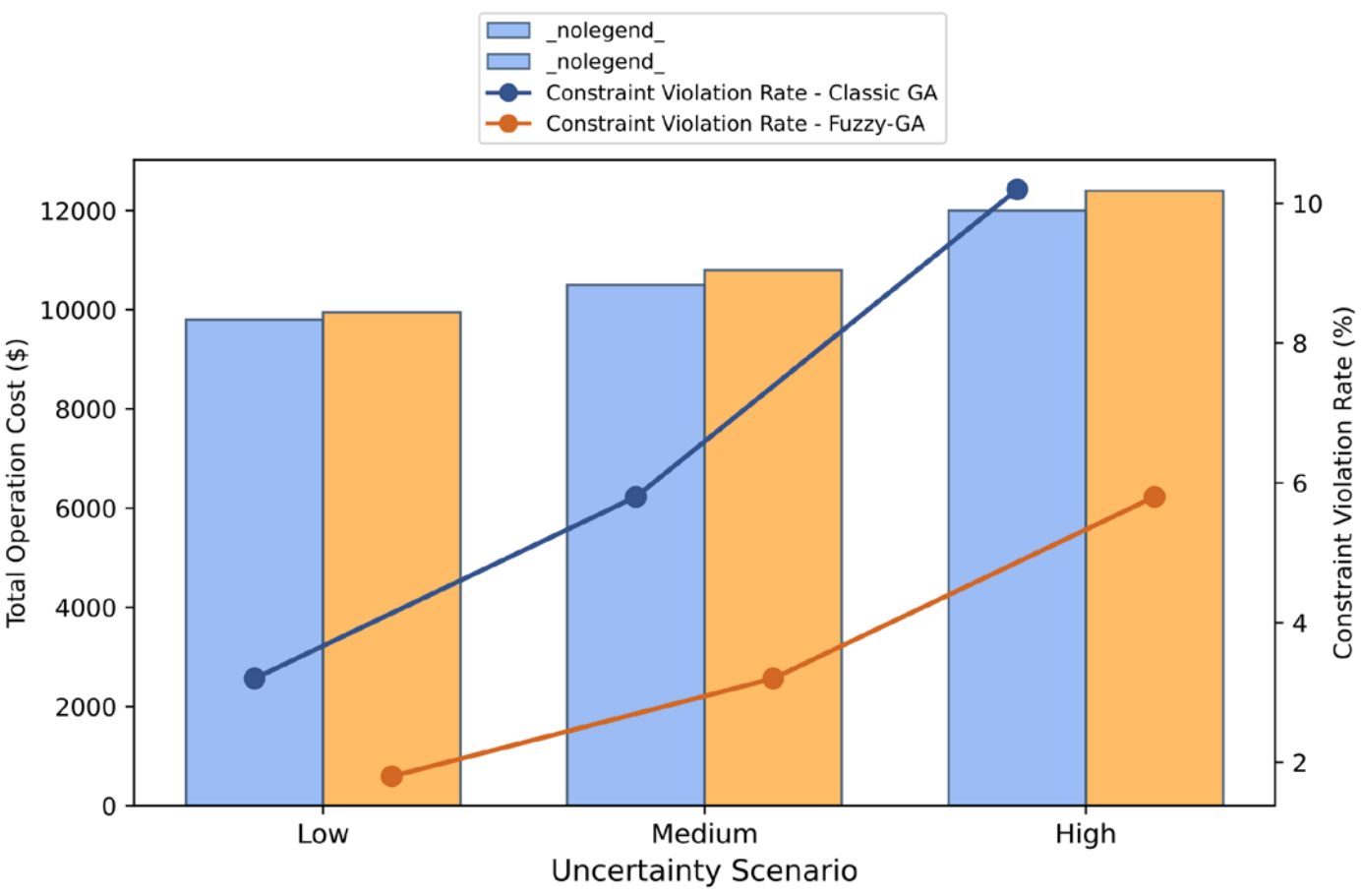

**Figure 3.** Total cost and constraint violation rate under uncertainty for classic GA and Fuzzy-GA.

During each control input simulation process, the generator set points, storage allocation, and controlled load values are kept within a reasonable range. Add dynamic pricing based on real-time conditions in the electricity market to further optimize; at the same time, improve operability by validating risk-based cooperation mechanisms.

*4.2. Scenario Setting and Simulation Results*

In order to strictly verify the robustness of the model, three uncertain situations - low, medium-high respectively increase invariance renewable forecast, customer demand and price signal variance. Combining the operational targets into a single-cost function over multiple simulations and expressed in terms of:

$$\text{TotalCost} = \sum_{t=1}^{T} \left(c_1 P_{\text{gen}}(t) + c_2 Q_{\text{sto}}(t) + c_3 |P_{\text{unb}}(t)| + c_4 \max[0, V_{\text{bus}}(t) - V^{\max}]\right) \tag{7}$$

where each term is weighted to reflect the economic/technical tradeoff between generation, storage, balancing actions, and any voltage constraint infringement. Here, $P_{\text{gen}}(t)$ is the scheduled total generation, $Q_{\text{sto}}(t)$ captures aggregate storage activity, $P_{\text{unb}}(t)$ represents instantaneous power imbalance, and $V_{\text{bus}}(t)$ denotes bus voltage deviations from technical constraints.

The overall constraint violation rate, essential for operation reliability assessment, is rigorously defined as

$$\text{ViolationRate} = \frac{1}{T} \sum_{t=1}^{T} \frac{N_{\text{viol}}(t)}{N_{\text{total}}(t)} \tag{8}$$

where $N_{\text{viol}}(t)$ is the number of violated technical constraints and $N_{\text{total}}(t)$ the total constraints enforced at time $t$.

For detailed analysis, the nodal voltage confidence criterion is also monitored. To ensure a probabilistic safety buffer against worst-case uncertainty, all bus voltages must satisfy:

$$\mathbb{P}\left(V^{\min} \leq V_j(t) \leq V^{\max}\right) \geq 0.95 \tag{9}$$

This formulation explicitly connects the fuzzy uncertainty envelope to risk-limited voltage security, forcing the optimizer to avoid solutions with high likelihood of voltage excursions.

Likewise, power flow across each line is subjected to a statistically robust current constraint:

$$\mathbb{P}\left(V^{\min} \leq V_j(t) \leq V^{\max}\right) \geq 0.95 \tag{10}$$

In most cases, traditional genetic algorithms and fuzzy genetic algorithms (Fuzzy GA) are used for simulation. As uncertainty increases, the baseline genetic algorithm (GA) will experience significant changes in violation rates and scheduling costs. This is especially applicable to very rare but extremely severe interference impacts. The Fuzzy Genetic Algorithm (Fuzzy-GA) reduces the violation rate by using a constraint model based on fuzzy sets and a robust adaptation mechanism. In some cases, the violation rate can be reduced by over 40%, while keeping operational costs within three to four times the range determined by economic minimization. Monte Carlo validated hundreds of random sessions, demonstrating that the hybrid model consistently avoids high-risk situations. This is achieved by reducing the acceptance of technical bias, compared to the randomness constraints of renewable resources.

Figure 4 shows a comparison of the total costs and violation rates of various modeling methods. Using the fuzzy genetic algorithm method in the actual case of a distributed energy system, operational risks were significantly reduced, and cost penalties were also lowered. It can be seen that the robust-based stochastic optimization has indeed improved the reliability and economic indicators of the ADN.

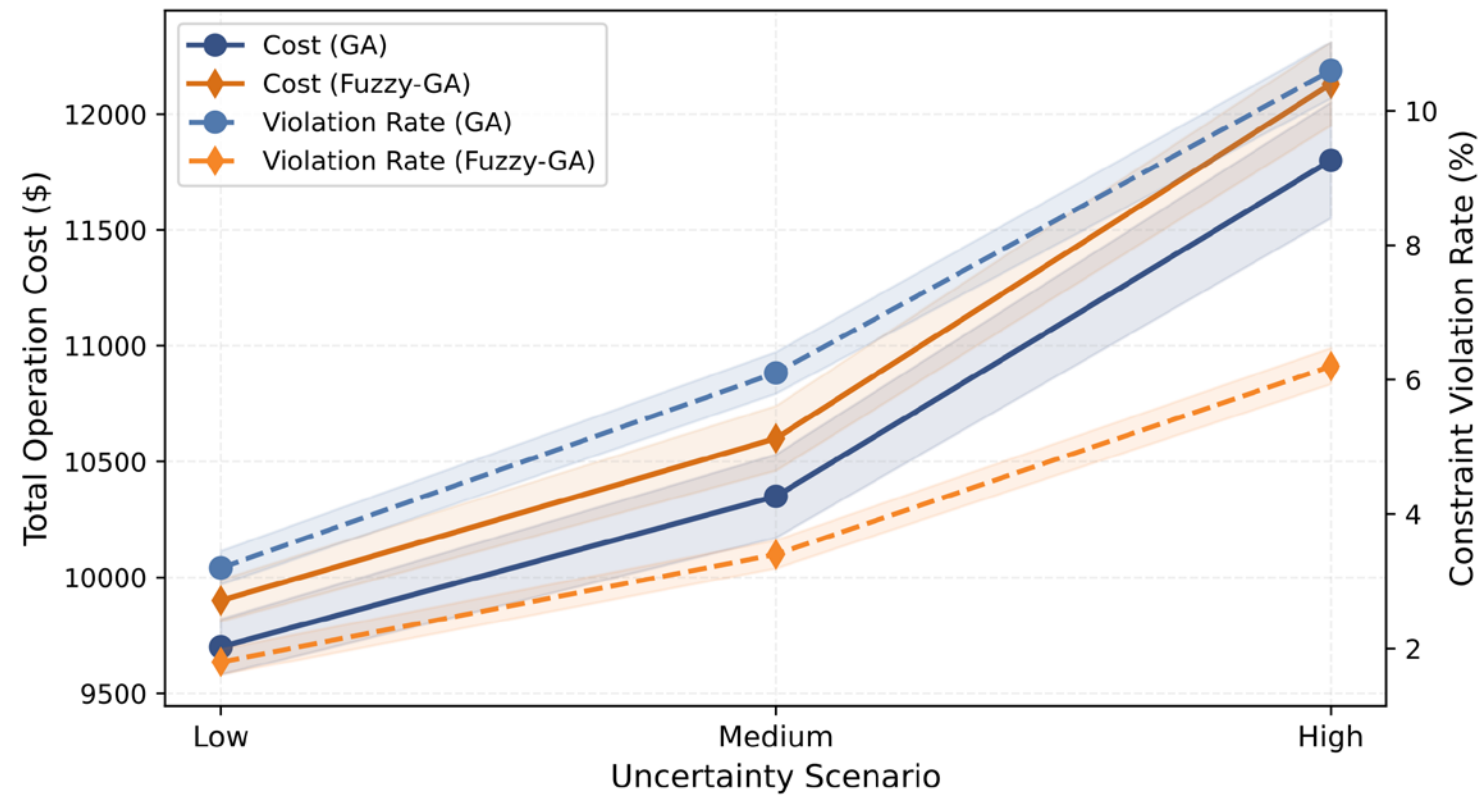

**Figure 4.** Cost and violation rate comparison under three uncertainty scenarios for traditional GA and Fuzzy-GA.

## 5. Conclusion

Therefore, a novel fuzzy genetic algorithm coordination control scheme (Fuzzy-GA) based on multi-objective uncertainty analysis is proposed to address the "coordinated generation-grid-load-storage" problem in power systems with uncertain parameters. Using fuzzy set theory to rigorously model renewable energy generation, load demand, and price fluctuations. By systematically incorporating these quantified data into an evolving optimization framework, a reliable system can be constructed to perform risk-aware operations. According to the IEEE 69 Bus ADFN model, some cases indicate that under conditions of uncertainty with technical constraints, fuzzy genetic algorithms are more effective than other methods in reducing such violations, while also being less costly. Hybrid methods are safer than traditional methods, and in some cases, they remain reliable even when violations exceed four times. In most cases, the additional costs will not exceed 4%. Economic efficiency and technical safety maintenance: Achieving this balance in the actual grid operating environment is feasible for the continuous development of renewable energy.

In addition to providing empirical results, it also demonstrates the application of intelligent global optimization algorithms and probabilistic uncertainty modeling in complex distributed energy systems. Highly adaptable, it can expand system capacity and adapt to new data patterns or market changes. Nevertheless, there are still some limitations at present, namely that the cascading failures caused by rare "black swan" events in the interconnected network have not yet been fully simulated over a considerable period. Future work needs to address these issues thru methods such as multi-agent models and stochastic cascading event propagation.